\documentclass[conference,authordraft,anonymous]{IEEEtran}
\IEEEoverridecommandlockouts
\usepackage{cite}
\usepackage{colortbl}
\usepackage{amsmath,amssymb,amsfonts}

\usepackage{algorithm}
\usepackage{algpseudocode} 
\usepackage{booktabs}
\usepackage{multirow}
\usepackage{graphicx}
\usepackage{textcomp}
\usepackage{xcolor}
\usepackage{color}
\usepackage{float}      
\usepackage{subfig}          
\newcommand{\figref}[1]{\figurename~{\ref{#1}}}

\newcommand{\secref}[1]{Section~{\ref{#1}}}

\def\BibTeX{{\rm B\kern-.05em{\sc i\kern-.025em b}\kern-.08em
    T\kern-.1667em\lower.7ex\hbox{E}\kern-.125emX}}
\begin{document}

\title{DyBit: \underline{Dy}namic \underline{Bit}-Precision Numbers for Efficient Quantized Neural Network Inference}




\author{
\IEEEauthorblockN{Jiajun Zhou$^{1*}$, Jiajun Wu$^{1*}$, Yizhao Gao$^{1}$, Yuhao Ding$^{1}$, Chaofan Tao$^{1}$, Boyu Li$^{1}$ \\ Fengbin Tu$^{2}$, Kwang-Ting Cheng$^{2}$, Hayden Kwok-Hay So$^{1}$, Ngai Wong$^{1\dag}$} 
\IEEEauthorblockA{$^{1}$ Department of Electrical and Electronic Engineering,
The University of Hong Kong, Hong Kong} 
\IEEEauthorblockA{$^{2}$ Department of Electronic and Computer Engineering, The Hong Kong University of Science and Technology, Hong Kong} 
\IEEEauthorblockA{\{jjzhou, jjwu, yzgao, yhding\}@eee.hku.hk, \{cftao, liboyu\}@connect.hku.hk \\ tufengbin@gmail.com, timcheng@ust.hk, \{hso, nwong\}@eee.hku.hk 
}
 
}

\maketitle

\begingroup\renewcommand\thefootnote{*}
\footnotetext{Both authors contributed equally to this research}
\endgroup

\begingroup\renewcommand\thefootnote{\dag}
\footnotetext{Corresponding author}
\endgroup

\begin{abstract}

To accelerate the inference of deep neural networks (DNNs), quantization with low-bitwidth numbers is actively researched. A prominent challenge is to quantize the DNN models into low-bitwidth numbers without significant accuracy degradation, especially at very low bitwidths ($<$ 8 bits). This work targets an adaptive data representation with variable-length encoding called DyBit. DyBit can dynamically adjust the precision and range of separate bit-field to be adapted to the DNN weights/activations distribution. We also propose a hardware-aware quantization framework with a mixed-precision accelerator to trade-off the inference accuracy and speedup. Experimental results demonstrate that the inference accuracy via DyBit is 1.997\% higher than the state-of-the-art at 4-bit quantization, and the proposed framework can achieve up to 8.1$\times$ speedup compared with the original model.
\end{abstract}

\begin{IEEEkeywords}
Deep Neural Networks, Quantization, Accelerator
\end{IEEEkeywords}

\section{Introduction}

There is an ever-growing need of accelerating deep neural network (DNN) inference.
While the de facto industrial standard is to represent network weights as single-precision ($32$-bit) floating-point (FP) numbers in pre-trained DNN models, inference hardware commonly relies on reduced bitwidth fixed point arithmetic circuits (e.g., \texttt{int8}) instead for their superior speed, area, and energy efficiency over their floating-point counterparts.
To operate with these fixed point hardware, the original FP models must first be quantized into the target low-precision linear fixed-point representations offline based on the training data~\cite{jacob2018quantization}.
Although the use of low-bitwidth hardware can significantly speed up DNN inference, this approach suffers from significant accuracy degradation, especially on very low bitwidth settings ($< 8$ bits), because it is challenging for the \emph{fixed} and \emph{linear} range of the conventional fixed-point format to capture the complex \emph{dynamic} parameter distribution changes in a DNN model during run time.
A number of recent works attempted to address this challenge by introducing mixed-precision quantization that employs fixed-point numbers of different bitwidth in different parts of the neural network~\cite{wang2019haq, li2021brecq, hao}.
Unfortunately, obtaining the optimal configuration for mixed-precision quantization that minimizes accuracy loss remains an unsolved problem, making it difficult to justify the hardware and speed overhead of supporting mixed-precision operations in hardware~\cite{wang2019haq}.
\begin{figure}[tbp]
\setlength{\belowcaptionskip}{-0.3cm}
\centering
\includegraphics[scale=0.9]{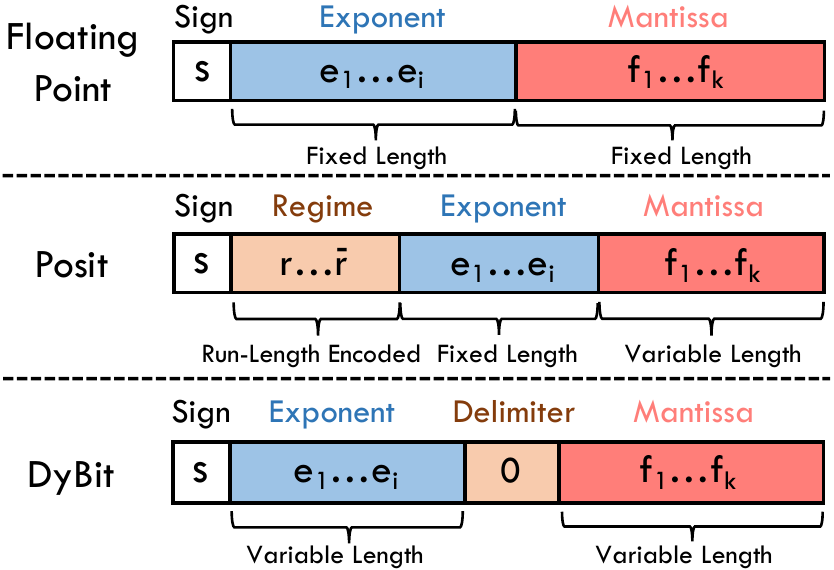}
\caption{An illustration of different $N$-bit numerical arithmetic formats including  FP, Posits and DyBit numbers.}
\label{fig:dac_fig1}
\vspace{-1mm}
\end{figure}
To address the low-bitwidth quantization challenge caused by the linear mapping in conventional fixed-point representations, recent works have begun to investigate in tailored made number presentations for neural network inference that reduce the representation error of low-bitwidth quantization~\cite{langroudi2021alps, adaptivfloat, guo2022ant}. Instead of affine mapping, they leverage additional mechanisms to adjust precision in different ranges. For instance, Posit~\cite{langroudi2021alps} uses run-length encoding to dynamically define the exponent and mantissa ranges in each data, while Adaptivfloat \cite{adaptivfloat} assigns different exponent lengths to different data blocks as an adjustment for precision ranges. These approaches are often designed in a way that can better represent the network based on their distribution properties due to the dynamic precision range with lower bitwidths than the standard FP format. However, existing adaptive data types require additional variables for adjusting the dynamic range. In this regard, a hardware-efficient data format that can dynamically represent the tensor distribution without extra variables is of research and practical value.

To this end, we propose a hardware-efficient data representation called \textit{DyBit} for low-bitwidth quantization with a variable length in the exponent bit-field to adapt to the distribution of DNN models. Furthermore, an efficient mixed-precision quantization framework is developed to tradeoff between quantization error and latency speedup. Thanks to the dynamically adaptive representation, the framework can quantize activations and weights to the lowest 4 bits and 2 bits, respectively, while maintaining high accuracy. The proposed framework can also be adapted to different application requirements using different constraints on quantization error or speedup. Finally, we design and implement a run-time configurable mixed-precision accelerator that can efficiently decode the DyBit and reuse computation units for different bitwidths. The key contributions of this work are:

\begin{itemize}
\item We propose \textit{DyBit}, an adaptive data representation that has efficient variable-length exponent bits and can also adjust its precision at the tensor level. Evaluation results show the proposed representation can be adapted to the data distributions in various DNN models and layers.

\item We have developed a run-time configurable mixed-precision accelerator that supports \textit{DyBit} operations, which fuses multiple multiply-accumulate (MAC) operations into one processing element to speed up the DNN inference and reduce memory access in low-bitwidth quantization.

\item We propose a hardware-aware mixed-precision quantization framework based on the adaptive \textit{DyBit} to trade-off between the inference accuracy and hardware speedup. The proposed framework searches for optimal layer-wise quantization based on two strategies for different application scenarios. 
\end{itemize}

\section{BACKGROUND AND RELATED WORK}
\subsection{Quantization Method}
Many studies have extensively explored DNN compression and optimization on hardware using quantization. For efficient edge deployment, binary neural networks (BNNs) exclusively make use of the logical XNOR operation that obviates regular multipliers binarized the network weights into \{-1,+1\}~\cite{rastegari2016xnor} and replace multiplication with addition or bit-shift operations. Jacob~\cite{jacob2018quantization} made use of fixed-length integers to quantize weights and activations. Many approaches only quantize static weights with on-device storage considerations but do not deliver verifiable computational efficiency improvements on real hardware. The survey paper by Qualcomm AI research~\cite{krishnamoorthi2018quantizing} contains more details about hardware-motivated methods for quantization. Nonetheless, these conventional quantization methods simply assign separate quantizers per group of weights and activations, whereas the proposed framework herein automates fused multiple bits for efficient calculations.

\subsection{Mixed-Precision Hardware Accelerator}
To efficiently support mixed-precision quantization, previous works have explored different architectural designs that can achieve scalable performances on different precisions. Prior mixed-precision accelerators can mainly be divided into \emph{spatial-based} and \emph{temporal-based} architectures depending on how the precision-scaling operations are mapped \cite{Review_mp}. The spatial-based accelerators, e.g., BitFusion \cite{sharma2018bit}, are generally based on a configurable multiplier composed of low-bitwidth multiply units. By splitting the full-precision input into several low-precision data, the multiplier can process multiplications with different bitwidths. On the other hand, temporal-based accelerators usually leverage bit-serial MAC operations to achieve efficient computation~\cite{bitcluster}. In general, it is challenging to achieve good tradeoffs between bit-level compatibility, energy efficiency, and performance without causing significant overhead to support mixed-precision operations. In this work, an accelerator based on a spatial-based architecture is proposed to efficiently support DyBit format under low precision ($<$8bits). A cycle-accurate simulator is also developed to foster hardware-aware mixed-precision quantization. 

\section{METHODOLOGY}

We now present more details of the proposed DyBit representation and the efficient hardware accelerator, as well as the mixed-precision framework and the quantization algorithm.

\begin{figure}[tbp]
    \centering
    \includegraphics[scale=0.57]{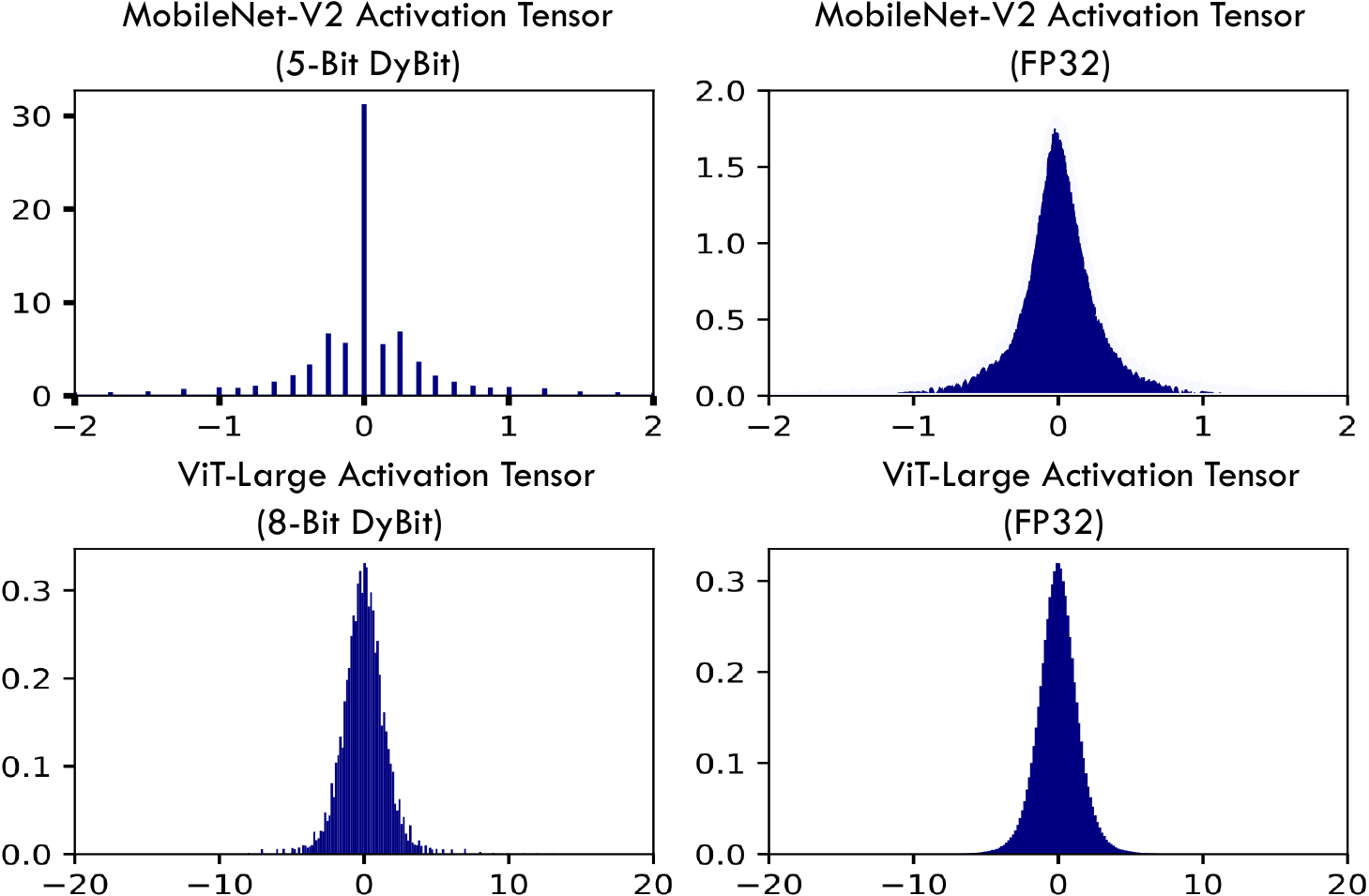}
    \caption{The diagram of the proposed DyBit quantization.}
    \label{fig:Dybit_quant}
\end{figure}

\begin{figure*}[tbp]
    \setlength{\belowcaptionskip}{-0.4cm}
    \centering
    \begin{minipage}[b]{0.475\textwidth}
            \centering
            \subfloat[]{\includegraphics[width=1\linewidth]{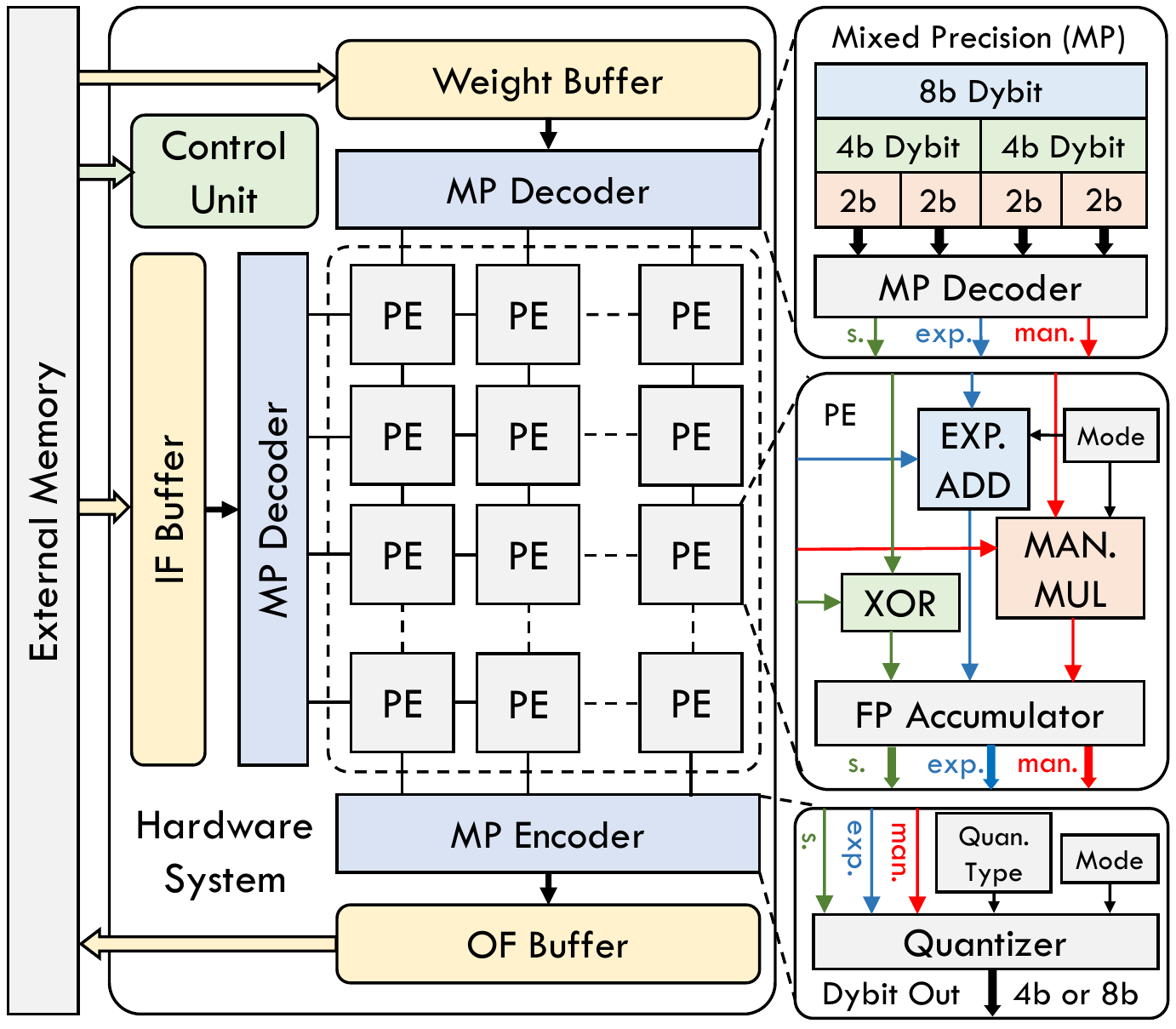}\label{subfig:hw-arch}}
    \end{minipage}
    \begin{minipage}[b]{0.517\textwidth}
            \centering
            \subfloat[]{\includegraphics[width=1\linewidth]{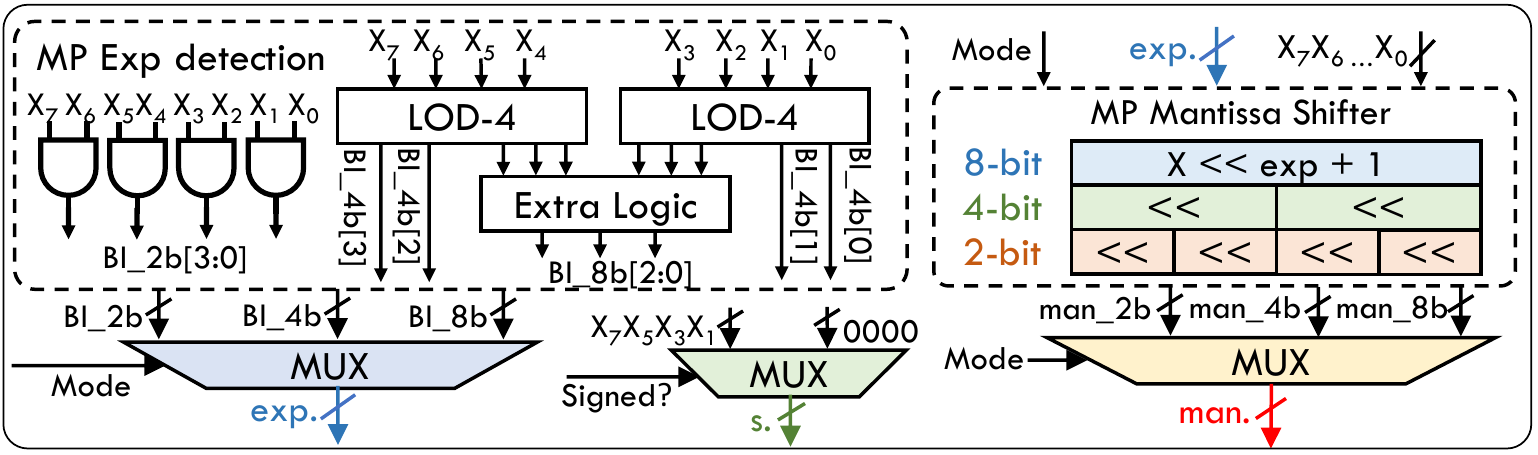}\label{subfig:hw-dec}}
            \
            \subfloat[]{\includegraphics[width=1\linewidth]{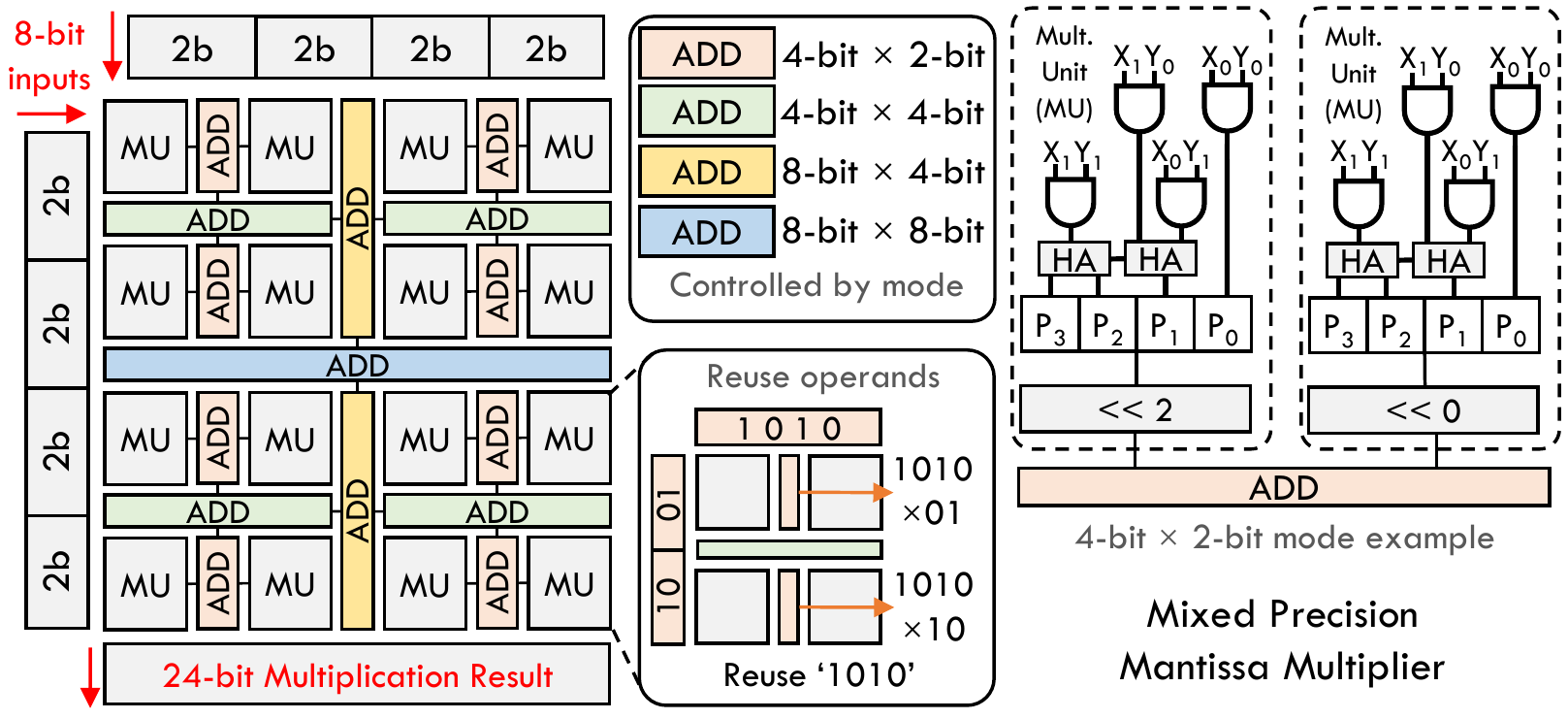}\label{subfig:hw-man-mul}}
    \end{minipage}
    \caption{Mixed-precision hardware system based on the proposed DyBit representation. (a) Hardware architecture based on the systolic array, (b) mixed-precision decoder (MP Decoder), and (c) mixed-precision mantissa multiplier (MAN. MUL).}
\end{figure*}

\begin{table}[htbp]
\centering
\caption{4-Bit Unsigned DyBit Value Table}
\label{tab:true_table}
\resizebox{\columnwidth}{!}{%
\begin{tabular}{|c|c|c|c|c|c|c|c|}
\hline
\textbf{Binary} & \textbf{Value} & \textbf{Binary} & \textbf{Value} & \textbf{Binary} & \textbf{Value} & \textbf{Binary} & \textbf{Value} \\ \hline
0 0 0 0 & 0     & 0 1 0 0 & 0.5   & 1 0 0 0 & 1.0  & 1 1 0 0 & 2 \\ \hline
0 0 0 1 & 0.125 & 0 1 0 1 & 0.625 & 1 0 0 1 & 1.25 & 1 1 0 1 & 3 \\ \hline
0 0 1 0 & 0.25  & 0 1 1 0 & 0.75  & 1 0 1 0 & 1.5  & 1 1 1 0 & 4 \\ \hline
0 0 1 1 & 0.375 & 0 1 1 1 & 0.875 & 1 0 1 1 & 1.75 & 1 1 1 1 & 8 \\ \hline
\end{tabular}%
}
\end{table}
\subsection{Variable-Length Datatype}
The variable-length DyBit number representation scheme contains a mandatory sign, multiple dynamical exponent bits, and mantissa bits. We follow the generic representation to illustrate any DyBit value in Fig.~\ref{fig:dac_fig1}. To efficiently decode all bitwidth data points, only shift or add operations are required to compute the bit-level number system. Specifically, the tapered DyBit representation is defined in Eqn.~(\ref{equation:dybit}).
\begin{align}
\footnotesize
\label{equation:dybit}
\mathbf{{\normalsize  f(x) = \begin{cases}
0 , &0  \\
2^{n} , &max  \\
{ (-1)^{s} \times\  2^{i-1} \times \left ( 1 + \frac{x}{2^{k}} \right )}, &others\\
\end{cases}} } 
\end{align}
where $n$ refers to the total number of bits, $i$ stands for the variable length of exponent bits, $k$ is the variable length of the power-of-2 scaling mantissa bits of encoded variable-length range bits, and $x$ represents the decimal number of the fraction field. In the formula, when the start bit is the digit 1, the variable-length exponent bit $i$ is used to encode the number of 1s and combines the hardware-oriented characteristics of leading one detector (LOD), which counts the number of 1s before the next zero bit. If the start bit is the digit 0, only variable-length fraction bit $k$ represents the actual value within \{-1,1\}. In this way, the exponent region of DyBit is a variable-length encoding method instead of a fixed-length one. Meanwhile, the fraction bits are also adaptively changed due to the shifting of the exponent bit. We further explain this encoding results of non-uniform distributions with a 4-bit truth Table~\ref{tab:true_table} that maps small and large values to tensor distributions. Thanks to the variable-length method, DyBit is suitable for DNN quantization as it can be adapted to tensor distributions of the original models (cf.~\figref{fig:Dybit_quant}).

\subsection{Hardware Design}\label{subsec:hw-design}

To support the DyBit-based quantization and inference, we propose a run-time configurable mixed-precision hardware accelerator. This section introduces how the architecture and circuit design efficiently support the configurable mixed-precision requirement.


\subsubsection{Architecture} The proposed hardware architecture is based on a systolic array with an input feature (IF) buffer, a weight buffer, and an output feature (OF) buffer, shown in \figref{subfig:hw-arch}. Based on the systolic dataflow, all partial results can remain FP for MAC operations. Thereby, all processing elements (PE) share the same decoder per row/column and the same encoder per column so that the decoders and encoders do not exist in PEs, which reduces the hardware overhead. The FP intermediate results will be quantized to DyBit format before being written back to the external memory.

\subsubsection{Decoder \& Encoder} Due to the mixed-precision support, decoding the input data into unified floating-point formats will be easier for processing. As in \figref{subfig:hw-dec}, the proposed mixed-precision decoder extracts the exponent ($exp$) by detecting the number of the leading 1s. Then the decoder left-shifts $exp$ to get the mantissa and inserts the normalized 1 in the MSB. Take an unsigned 8-bit DyBit data \texttt{11001010} as an example, the decoded data will be exponent(\texttt{001}), mantissa(\texttt{10101000}). To reduce the mixed-precision overhead, we reuse the 4-bit leading one detector (LOD-4) for 8-bit DyBit input, and we also reuse the logic in the dynamic shifter for the mantissa. For the encoder part, the process is the opposite of the decoding part, in which the circuit will insert $(exp+1)$ number of 1s in the MSB and select the remaining bits of mantissa to fill the DyBit output.

\subsubsection{Mixed-precision PE} As illustrated before, the data processed inside PEs fit well with variable-length separate bit-field. Implementing individual exponent adders and mantissa multipliers for different data widths will cause huge overhead as no computation resources are reused. For the mantissa multiplier (MAN. MUL), we modified the BitFusion~\cite{sharma2018bit} architecture to support four different multiplication modes. It is worth noting that based on this fused strategy, the PE can process multiple multiplications in parallel with data reuse (cf. \figref{subfig:hw-man-mul}). For the exponent adder (EXP. ADD), it is natural and trivial to reuse the low-precision adder to build up a high-precision adder with a small amount of overhead in the carry chain. The run-time instructions can control the PE working on different modes. With such mixed-precision PEs, when an $N \times N$ systolic array is working on $P_{1} \times P_{2}$ ($<$ 8-bit) mode, it is equivalent to achieving $(8/P_{1})N \times (8/P_{2})N$ scale based on this fixed systolic array. Therefore, our hardware design can achieve high speedup in low-precision modes.

\subsection{Hardware-aware Quantization Framework}\label{subsec:quant-framework}
\begin{figure}[tbp]
    \setlength{\belowcaptionskip}{-0.4cm}
    \centering
    \includegraphics[width=0.9\linewidth]{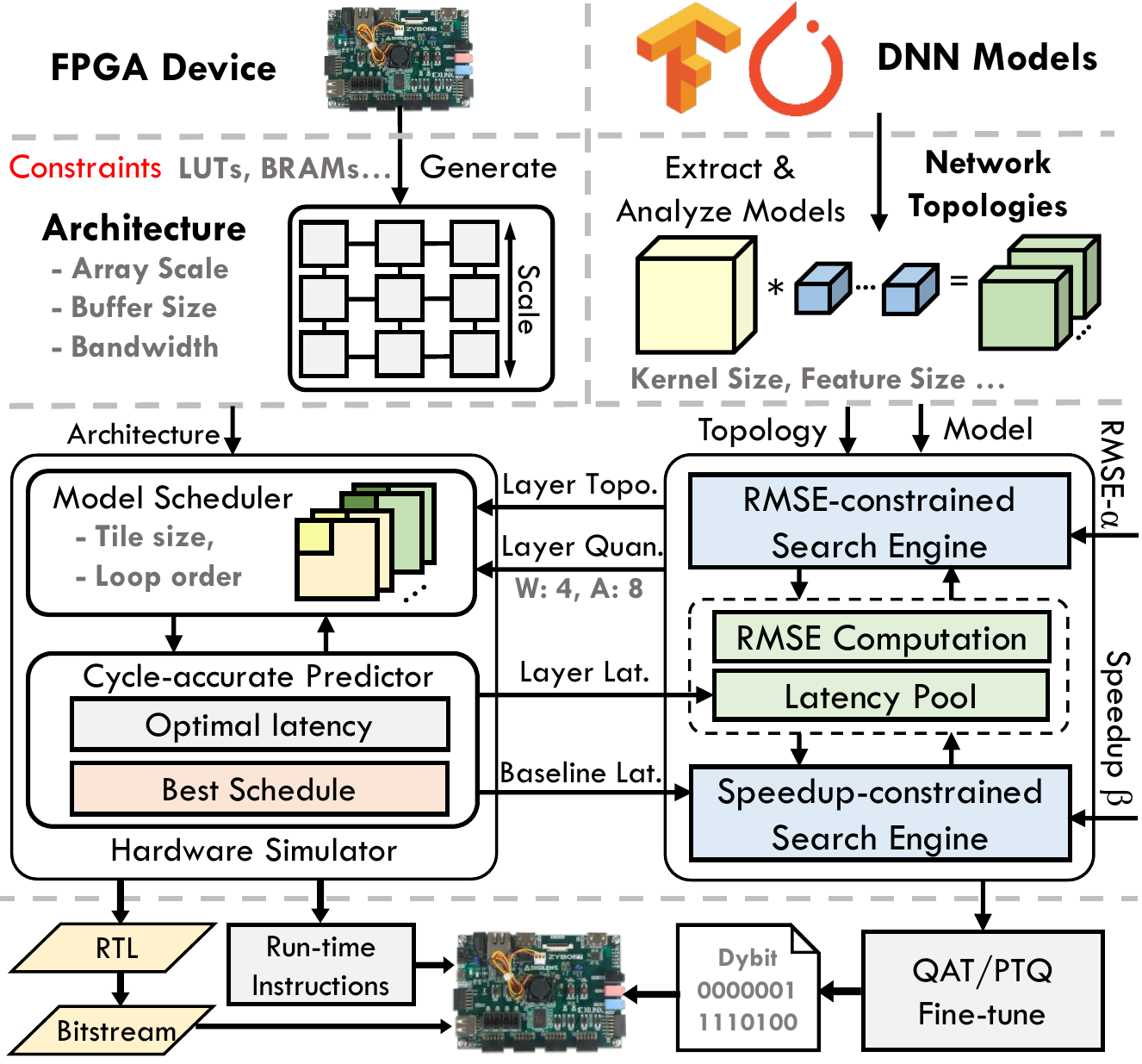}
    \caption{DyBit-Based hardware-aware quantization framework.}
    \label{fig:framework}
\end{figure}

\figref{fig:framework} presents the proposed novel hardware-aware quantization framework based on the search-based method. The framework first estimates the maximum hardware resource utilization based on the DNN models and given hardware constraints (e.g., LUTs and BRAMs in FPGAs). Then, it searches the layer-wise quantization bitwidths based on two variant-constrained strategies. The hardware-aware framework uses a cycle-accurate hardware simulator to provide latency results to do layer-wise mixed-precision quantization dynamically. Finally, the pre-trained 32-bit floating-point (FP32) models are quantized into DyBit according to the layer-wise search results using quantization-aware training (QAT) to retain accuracy. The post-quantization DNN models can then be deployed to our hardware accelerator.


\subsubsection{Quantization Metrics} According to previous works, Root Mean Squared Error (RMSE) is a common metric to effectively evaluate the accuracy of the post-quantization DNN models~\cite{adaptivfloat}. The smaller the RMSE, the higher accuracy a quantized model can potentially achieve. Here we use RMSE as a metric to measure the quantization error and facilitate the search process, defined as:
\begin{align}
\footnotesize
 RMSE = \sqrt{\frac{1}{n}\Sigma_{i=1}^{n}{\Big(\frac{x -\hat{x}}{\sigma_i}\Big)^2} },
    \label{eq:rmse}
\end{align}
$x$ and $\hat{x}$ are respectively the original FP32 and quantized values, and $\sigma_i$ is the standard deviation of the tensor distribution. 

\subsubsection{Two Search Strategies} 
Based on the quantization error RMSE and speedup ratio, we propose two different optimization strategies to adapt to different application scenarios. In the case of stringent real-time requirements, our framework can constrain the speedup ratio as $\alpha$ to ensure the hardware performance while minimizing the quantization error RMSE, as shown in Eqn.~(\ref{eq:lat-cons}). 
\begin{align}
\footnotesize
    \begin{split}
        \min_{\mathbf{A}}\quad  & \sum_{i = 1}^{m} RMSE_{i}(a,w) \\
        \text{s.t.}\quad        & \alpha\times\sum_{i = 1}^{m} Lat_{i}(a,w) \le \sum_{i = 1}^{m} Lat_{i}(8,8). \\
    \end{split}
    \label{eq:lat-cons}
\end{align}
On the other hand, if the application prioritizes accuracy, our framework can use another constraint $\beta$ to limit quantization error while obtaining quantization with minimum latency, as in Eqn.~(\ref{eq:rmse-cons}). Note that in both search strategies of our algorithm, we select 8-bit Dybit as the baseline for latency and RMSE metrics.
\begin{align}
\footnotesize
    \begin{split}
        \min_{\mathbf{A}}\quad  & \sum_{i = 1}^{m} Lat_{i}(a,w) \\
        \text{s.t.}\quad        & \sum_{i = 1}^{m} RMSE_{i}(a,w) \le \beta \times \sum_{i = 1}^{m} RMSE_{i}(8,8) \\
    \end{split}
    \label{eq:rmse-cons}
\end{align}

\subsubsection{Quantization Search Flow}
The layer-wise mixed-precision quantization of weights and activations leads to a vast design space. In DyBit, we support the selections for weights/activations in 8-bit, 4-bit, and 2-bit for better hardware efficiency since the bitwidths non-integer powers of 2 (e.g., 6-bit) will cause additional overhead for data alignment in off-chip memory and data transfer between accelerator and memory. Assuming the DNN model has $N$ layers, the total number of possible solutions will be $(3\times3)^{N}$. In such a space, we design a heuristic search algorithm to find near-optimal solutions efficiently.


Algorithm \ref{alg:dual-mode-search} describes the proposed heuristic search algorithm. In the speedup-constrained strategy, we get the layer-wise baseline latency performances calculated by the simulator and select the $k$ largest layers as candidates. In other words, we intend to quantize the slowest layer first to get a better overall end-to-end speedup. Besides, to get the optimal solution with the minimum RMSE, we also calculate the RMSE of each candidate and reorder them in ascending order of RMSE. The search engine will lower the bitwidth of each candidate one by one so that the low-RMSE layers can be quantized first. Whenever the speedup ratio within the $k$ candidates is satisfied, this iteration will stop. The engine will recalculate the latency and select the next top-$k$ candidates in the next iteration. The overall process will stop when the end-to-end speedup constraint is satisfied. This way, we can ensure the final speedup ratio while lower the RMSE as well.

As for the RMSE-constrained strategy, the objective and condition are exchanged compared with the speedup-constrained one. Therefore, the search flow is similar to the speedup-constrained strategy, except the ordering of the candidate is based on different metrics. 


\begin{algorithm}[tbp]
\footnotesize
\renewcommand{\algorithmicrequire}{\textbf{Input:}}
\renewcommand{\algorithmicensure}{\textbf{Output:}}
\caption{Search Flow of speedup-constrained and RMSE-constrained strategies}
\begin{algorithmic}[1]
\Require DNN model $\mathbf{M}$ with $N$ layers $\{L_{1}, L_{2}, ...L_{N}\}$, search strategy $m$, constraint $\alpha$ or $\beta$, top-k parameter $k$ 
\Ensure Layer-wise bitwidths of weights and activations $\mathbf{W}=(W_{1}, W_{2}, ...W_{N}), \mathbf{A}=(A_{1}, A_{2}, ...A_{N})$
\State $\mathbf{W}, \mathbf{A} \gets (8, 8, ...,8)$ 
\State $metric\_base \gets \textsc{Total\_metric}(\mathbf{M}, \mathbf{W}, \mathbf{A}), ratio \gets 1$
\While{$ratio$ does not meet $\alpha$ or $\beta$}
    \State $\mathrm{metric} \gets \textsc{Layerwise\_metric}(\mathbf{M}, \mathbf{W}, \mathbf{A})$ 
    \If {$m = \mathrm{speedup}$}
        \State $\mathrm{metric\_top} \gets \textsc{Lat\_rank}(\mathrm{metric}, k)$
        \State $\mathrm{layer\_list} \gets \textsc{RMSE\_rerank}(\mathrm{metric\_top})$
    \ElsIf {$m = \mathrm{RMSE}$}
        \State $\mathrm{metric\_top} \gets \textsc{RMSE\_rank}(\mathrm{metric}, k)$
        \State $\mathrm{layer\_list} \gets \textsc{Lat\_rerank}(\mathrm{metric\_top})$
    \EndIf
    \State \textsc{Degrade\_level}($\mathrm{list}, \mathbf{W}$)
    \State \textsc{Degrade\_level}($\mathrm{list}, \mathbf{A}$)
\EndWhile
\\
\Procedure{Degrade\_level}{$\mathrm{layer\_list}, \mathbf{W}$ or $\mathbf{A}$}
    \For {$l = 1 \to k$}
        \State Degrade $W_{layer\_list\left[l\right]}$ or $A_{layer\_list\left[l\right]}$: 8 $\to$ 4 or 4 $\to$ 2
        \State $ratio \gets \textsc{Total\_metric}(\mathbf{M}, \mathbf{W}, \mathbf{A})/metric\_base$
        \State \textbf{break if} $ratio$ meets $\alpha$ or $\beta$
    \EndFor
\EndProcedure
\end{algorithmic}
\label{alg:dual-mode-search}
\end{algorithm}

\subsubsection{Hardware Simulator}
To support the hardware-aware quantization, we develop a cycle-accurate simulator, shown in \figref{fig:framework}. The simulator first generates the maximum architecture constrained by the resources of the target device. By modifying the backend of the systolic array GEMM dataflow~\cite{guo2022ant} based on our hardware design, it obtains the optimal latency by calculating the latencies corresponding to all possible tiling schedules of the current layer. During the search flow, the search engines call the simulator to get the latency of each layer, which will be used for ranking layer candidates.

\section{Evaluations}\label{sec:eval}

\subsection{Experiment Setup}
\subsubsection{Benchmark}
We conduct the experiments based on ResNet18/50 and the lightweight MobileNetV2 on ImageNet classification. We use the pre-trained 32-bit floating-point(FP32) model from PyTorch as the baseline. Based on the method in~\secref{subsec:quant-framework}, we train 3$\sim$5 fine-tuning epochs for QAT. To conduct a fair comparison, the training setup, and the hyper-parameters are kept the same for all types under evaluation. We also test our framework on emerging models like Vision Transformer, RegNet, and ConvNext to verify that method is universal and efficient.

\subsubsection{Baselines}
We obtained evaluation results for integer quantized models (i.e. INT4, INT8) following the same training setup. Besides, we compare our method with various fixed-precision quantization methods including PACT~\cite{choi2018pact}, AdaFloat~\cite{adaptivfloat}, DSQ~\cite{gong2019differentiable}, Posit~\cite{langroudi2021alps}, Flint~\cite{guo2022ant} and layer-wise mixed-precision quantization methods, such as BRECQ~\cite{li2021brecq}.
 
\subsubsection{Implementation}
The proposed mixed-precision accelerator is designed and implemented with Verilog HDL. We implemented the accelerator on the Xilinx ZCU102 platform. As discussed in~\secref{subsec:quant-framework}, a cycle-accurate hardware simulator is developed to support hardware-aware quantization search. We also utilize this simulator to evaluate our speedup performance compared with baselines.

\begin{table}[tbp]
\scriptsize
\caption{Top-1 accuracy performance with quantization-aware training on ImageNet dataset.}
\label{qat}
\resizebox{\linewidth}{!}{%
\begin{tabular}{@{}cccc@{}}
\toprule
\multicolumn{1}{c|}{Methods (W/A)} & \multicolumn{1}{c|}{MobileNetV2} & \multicolumn{1}{c|}{ResNet18} & ResNet50 \\ \midrule
FP32                & $71.79$ & $69.68$ & $75.98$ \\
INT(4/4)            & $39.78$ & $66.24$ & $73.04$ \\
INT(8/8)            & $71.658$ & $69.4$  & $75.92$ \\
AdaFloat(4/4) \cite{adaptivfloat}           & $-$ & $-$  & $75.1$ \\
BRECQ(4/4) \cite{li2021brecq}          & $66.57$ & $69.60$ & $-$     \\
PACT(4/4) \cite{choi2018pact}           & $61.40$ & $69.20$ & $-$     \\
DSQ(4/4) \cite{gong2019differentiable}            & $64.80$ & $69.56$ & $-$     \\
Flint(4/4) \cite{guo2022ant}          & $-$     & $67.50$ & $74.91$ \\ 
Posit(8/8) \cite{langroudi2021alps}         & $-$     & $-$      & $73.61$ \\ \midrule
\textbf{DyBit(4/4)} & $69.31$ & $69.47$ & $75.87$ \\
\textbf{DyBit(4/8)} & $68.17$ & $69.57$ & $75.82$ \\
\textbf{DyBit(8/8)} & $69.47$ & $69.66$ & $75.93$ \\ \bottomrule
\end{tabular}%
}
\end{table}

\begin{table}[t]
\scriptsize
\caption{Top-1 accuracy performance with quantization-aware training on ImageNet with emerging models}
\label{tab:qat_newmodels}
\resizebox{\columnwidth}{!}{%
\begin{tabular}{@{}cccc@{}}
\toprule
\multicolumn{1}{c|}{Methods (W/A)} & \multicolumn{1}{c|}{RegNet-3.2GF} & \multicolumn{1}{c|}{ConvNext-Tiny} & ViT-Base \\ \midrule
FP32                & 78.364 & 82.52 & 81.07 \\ 
INT(4/4)            & 75.9   & 0.1   & 72.19 \\
Flint(4/4) \cite{guo2022ant}         & -      & -     & 78.33 \\ \midrule
\textbf{DyBit(4/4)} & 77.13  & 71.9  & 79.44 \\
\textbf{DyBit(8/8)} & 77.844 & 80.55 & 80.82 \\ \bottomrule
\end{tabular}%
}
\vspace{-1mm}
\end{table}


\begin{figure*}
    \setlength{\belowcaptionskip}{-0.3cm}
    \centering
    \includegraphics[width=1\textwidth]{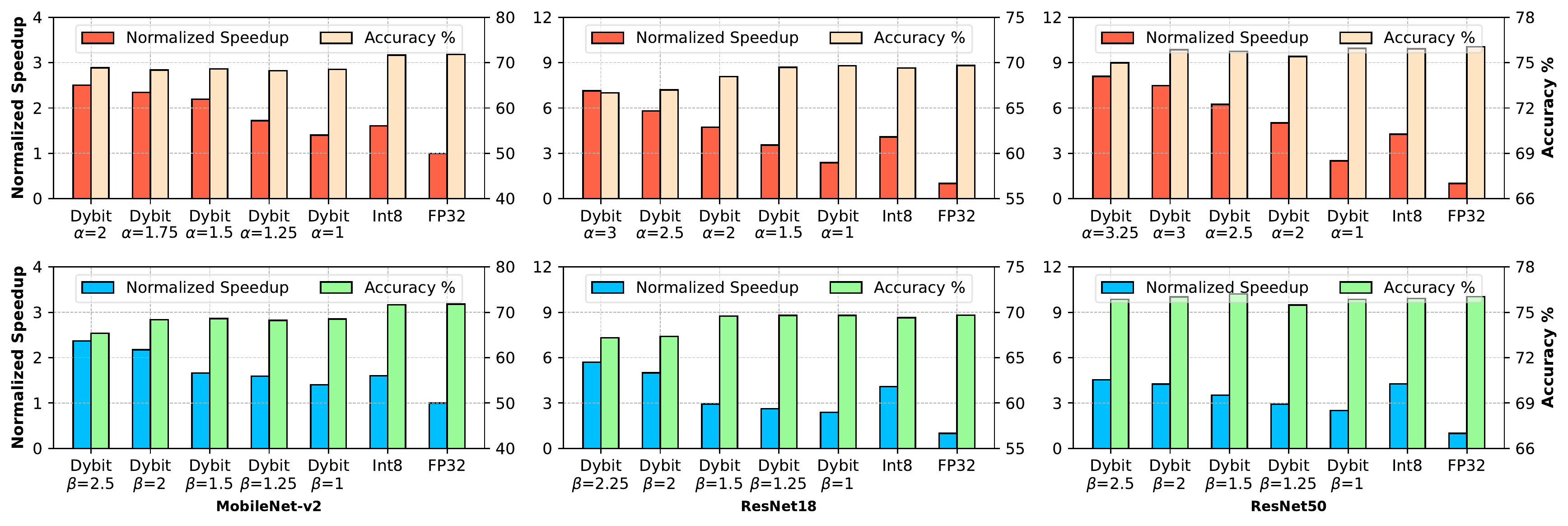}
    \caption{Speedup and accuracy evaluations on the speedup-constrained strategy (the first row) and the RMSE-constrained strategy (the second row), based on MobileNetV2 and ResNet18/50 models. The target platform is Xilinx ZCU102.}
    \label{fig:bar-figs}
\end{figure*}

\subsection{Quantization Accuracy}
To validate that the adaptive DyBit data representation can keep the accuracy in the low-precision models, we conducted quantization-aware training. In Table~\ref{qat}, we show the Top-1 accuracy results of three models in different bitwidth on the ImageNet dataset, e.g., 4W4A stands for 4-bit activation and 4-bit weight tensor. We observe that our quantization achieves 1.997\% inference accuracy higher than the state-of-the-art, viz. Flint~\cite{guo2022ant}, on 4-bit quantization and also surpasses other fixed-precision or mixed-precision quantization methods. Besides, it is noteworthy that 4-bit and 8-bit quantization results are provided in Table~\ref{tab:qat_newmodels} to demonstrate that our method for larger models causes less accuracy drop compared to high-precision models after fine-tuning. FP32, INT4, and INT8 results are also provided for a fair comparison. In addition, 8-bit DyBit has only a 0.05 Top-1 accuracy drop compared with FP32 on ResNet50. Specifically, our quantization method performs better at the lower-precision bitwidth.  

\begin{figure}[tbp]
    \setlength{\belowcaptionskip}{-0.3cm}
    \centering
    \includegraphics[width=1\linewidth]{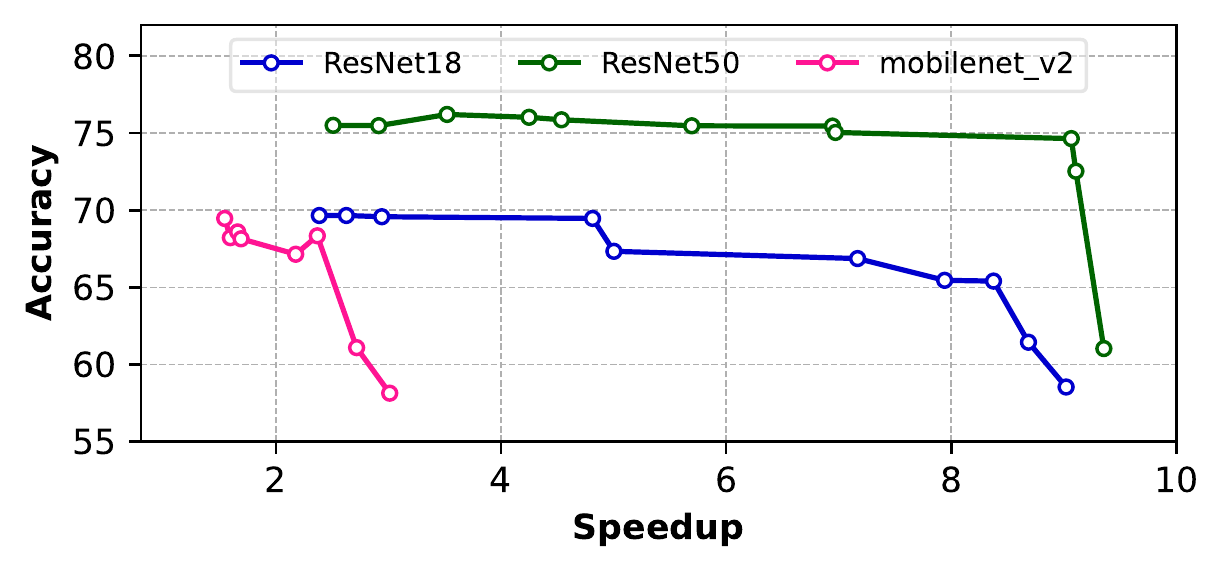}
    \caption{Accuracy-speedup trade-off in DyBit quantization.}
    \label{fig:acc-speed}
\end{figure}

\subsection{Accuracy-Speedup Trade-off}

To demonstrate the proposed hardware-aware quantization framework can trade-off between accuracy and speedup, we set up different constraints and quantize the ResNet18/50 and MobileNetV2 models based on the two search strategies in \secref{subsec:quant-framework}, depicted in \figref{fig:bar-figs}. Generally, an increase in the constraint $\alpha$ or $\beta$ leads to a speedup increase and accuracy loss, because the framework will search for more low-precision numbers to meet the demand. For the speedup-constrained strategy, the quantized model tends to have a higher speedup (e.g., in ResNet50, up to 8.1$\times$) with lower accuracy. On the contrary, the quantized model can maintain a closer accuracy to the original model while still delivering a decent speedup in the RMSE-constrained strategy (e.g., in ResNet50, only 0.18\% accuracy drop with 4.5 $\times$ speedup). Therefore, the proposed framework can work for different application scenarios with the two strategies. To further present the adjustment between accuracy and speedup, we collect all results based on both strategies, as shown in \figref{fig:acc-speed}. It can be concluded that with the growing speedup, the inference accuracy drops, and our proposed framework can quantize the DNN models with trade-offs along the curves. The speedup ratio is limited in the MobileNetV2 since depth-wise operations are not efficient based on our current GEMM systolic array.

\section{Conclusion}
This paper has proposed a novel hardware-aware quantization framework, with a fused mixed-precision accelerator, to efficiently support a distribution-adaptive data representation named DyBit. The variable-length bit-fields enable DyBit to adapt to the tensor distribution in DNNs. Evaluation results show that DyBit-based quantization at very low bitwidths ($<$8bits) consistently achieves higher accuracy than competing methods. Moreover, the proposed end-to-end framework can effectively search for the optimal solution under various constraints, thus achieving a trade-off between accuracy and hardware speedup. Experiments on various DNN models under different quantization constraints demonstrate that the framework can quantize DNN models to achieve $2.5 \sim 8.1\times$ speedup. 

\bibliographystyle{IEEEtran}
\bibliography{main.bib}


\end{document}